\documentclass[11pt,a4paper]{article}
\usepackage[hyperref]{acl2021}
\usepackage{times}
\usepackage{latexsym}

\usepackage{microtype}
\usepackage{booktabs}
\usepackage{multirow}
\usepackage{caption}
\usepackage{subcaption}
\usepackage{graphicx}

\aclfinalcopy 
\def\aclpaperid{3967} 

\title{Adapting Monolingual Models: \\Data can be Scarce when Language Similarity is High}

\author{Wietse de Vries\Thanks{ These authors contributed equally.} \And Martijn Bartelds\footnotemark[1] \And Malvina Nissim \and Martijn Wieling \vspace{5px}\\
 
        University of Groningen \\ The Netherlands \vspace{5px}\\
        \texttt{\{wietse.de.vries, m.bartelds, m.nissim, m.b.wieling\}@rug.nl}}

\date{}

\begin{document}
\maketitle
\begin{abstract}
For many (minority) languages, the resources needed to train large models are not available.
We investigate the performance of zero-shot transfer learning with as little data as possible, and the influence of language similarity in this process.
We retrain the lexical layers of four BERT-based models using data from two low-resource target language varieties, while the Transformer layers are independently fine-tuned on a POS-tagging task in the model's source language.
By combining the new lexical layers and fine-tuned Transformer layers, we achieve high task performance for both target languages.
With high language similarity, 10MB of data appears sufficient to achieve substantial monolingual transfer performance.
Monolingual BERT-based models generally achieve higher downstream task performance after retraining the lexical layer than multilingual BERT, even when the target language is included in the multilingual model.
\end{abstract}


\section{Introduction}
Large pre-trained language models are the dominant approach for solving many tasks in natural language processing.
These models represent linguistic structure on the basis of large corpora that exist for high-resource languages, such as English.
However, for the majority of the world's languages, these large corpora are not available.



Past work on multilingual learning has found that multilingual BERT~(mBERT; \citealt{devlin2019-mbert}) generalizes across languages with high zero-shot transfer performance on a variety of tasks~\citep{pires-etal-2019-multilingual, wu-dredze-2019-beto}.
However, it has also been observed that 
high-resource languages included in mBERT pre-training often have a better-performing monolingual model, and low-resource languages that are not included in mBERT pre-training usually show poor performance \citep{nozza2020mask, wu-dredze-2020-languages}.

An alternative to multilingual transfer learning is the adaptation of existing monolingual models to other languages.
\citet{zoph-etal-2016-transfer} introduce a method for transferring a pre-trained machine translation model to lower-resource languages by only fine-tuning the lexical layer.
This method has also been applied to BERT \citep{artetxe-etal-2020-cross} and GPT-2 \citep{devries2020good}.
\citet{artetxe-etal-2020-cross} also show that BERT models with retrained lexical layers perform well in downstream tasks, but comparatively high performance has only been demonstrated for languages for which at least 400MB of data is available.

To test if this procedure is also effective for low- to zero-resource languages, we consider two regional language varieties spoken in the North of the Netherlands, namely Gronings (Low Saxon language variant) and West Frisian.

\begin{figure}[hb!]
  \begin{center}
    \includegraphics[width=7.7cm]{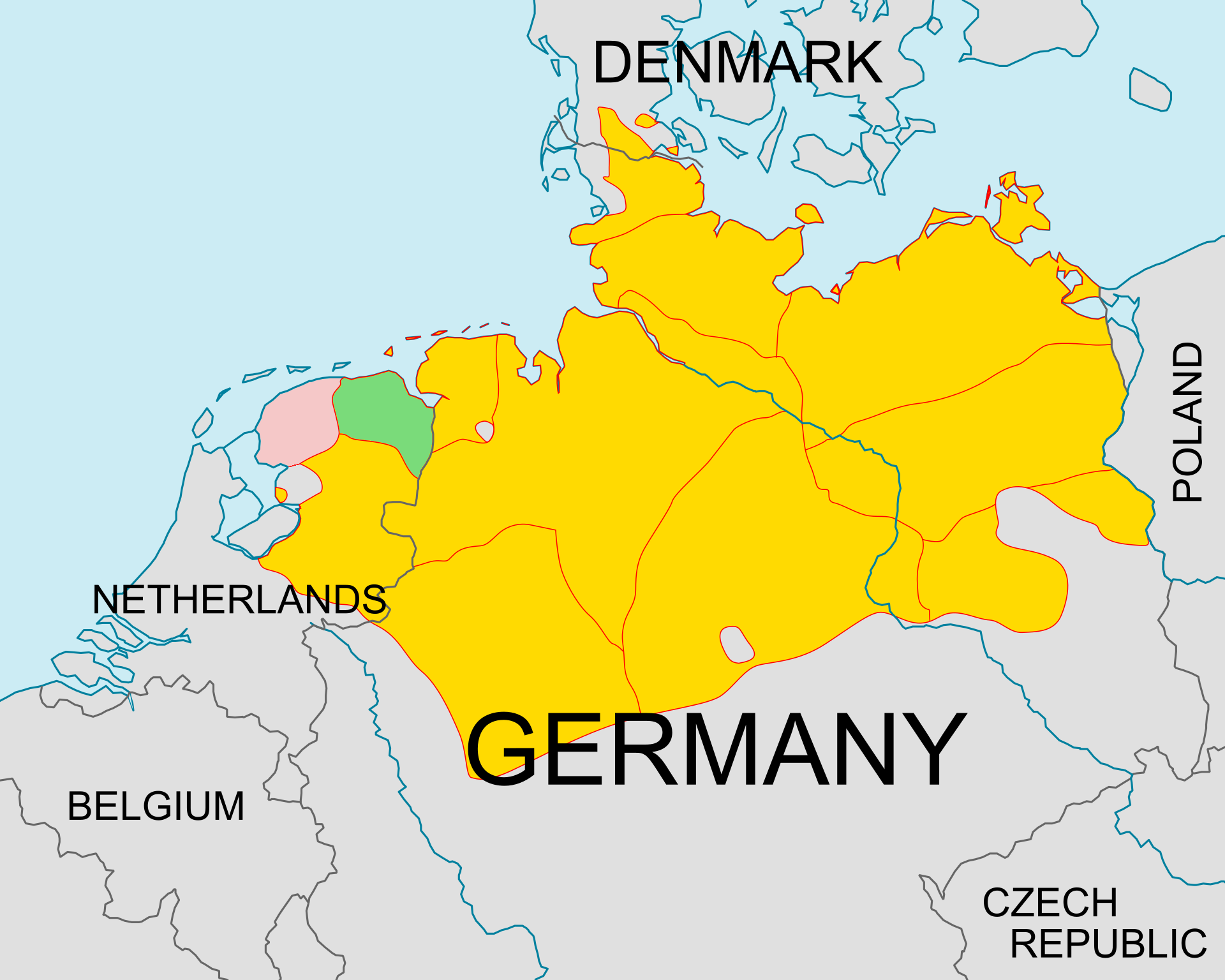}
    \caption{\label{appendix:fig:map} Geographical areas where Gronings (in green) and West Frisian (in red) are spoken. Image modified from \url{https://en.wikipedia.org/wiki/Low_German}.}
  \end{center}
\end{figure}

Figure~\ref{appendix:fig:map} visualizes the geographical areas where these regional language variants are spoken. The regional Low Saxon language is spoken in the north-eastern provinces of the Netherlands and in the North of Germany (shown in yellow). As part of the Low Saxon language, Gronings is spoken in the province of Groningen (highlighted in green). The West Frisian language is spoken in the province of Friesland (shown in red), and it is the second official language of the Netherlands, next to Dutch. Dutch is the national language of the Netherlands, and it is spoken in every province of the Netherlands and in Flanders (North of Belgium).

For both Gronings and West Frisian limited data is available. 
In addition to unlabeled data, for both target languages we have a small collection of annotated part-of-speech (POS) tagging data, which we use for evaluating zero-shot model transfer.
We use three monolingual BERT models (source languages English, German, Dutch) and mBERT to investigate if linguistic structure can be transferred to Gronings and West Frisian by learning new sub-word embeddings.
Our model source and target languages are closely related West Germanic languages \citep{eberhard_ethnologue_2020}.
In Table~\ref{tab:examples}, we show parallel sentences in Gronings, West Frisian, Dutch, German, and English to illustrate the lexical similarity between these languages. 
Additionally, the examples show that there are some lexical and syntactic differences.

\begin{table}[b!]
  \centering
  \resizebox{7.7cm}{!}{%
    \begin{tabular}{ll}
    \toprule
        Gronings     & Tom is n jong en Mary is n wicht. \\
        West Frisian & Tom is in jonge en Mary is in famke. \\
        Dutch        & Tom is een jongen en Mary is een meisje. \\
        German       & Tom ist ein Junge und Mary ist ein Mädchen. \\
        English      & Tom is a boy and Mary is a girl. \\
        \midrule
        Gronings     & Zie haar n bloum ien heur haand. \\
        West Frisian & Se hie in blom yn har hân. \\
        Dutch        & Ze had een bloem in haar hand. \\
        German       & Sie hatte eine Blume in der Hand. \\
        English      & She had a flower in her hand. \\
        \midrule
        Gronings     & Dat was n poar joar leden. \\
        West Frisian & Dat wie in pear jier lyn. \\
        Dutch        & Dat was een paar jaar geleden. \\
        German       & Das war vor ein paar Jahren. \\
        English      & That was a couple of years ago. \\
    \bottomrule
    \end{tabular}%
    }
  \caption{\label{tab:examples} Translations of three sentences in Gronings, West Frisian, Dutch, German, and English.}
\end{table}

We also evaluate to what extent the similarity between each source language of the monolingual models and the target languages is relevant for transferring monolingual representations, and assess the minimum amount of data necessary to adapt these models.

Our pre-trained models for Gronings and West Frisian (which did not yet exist) are released. 
Additionally, our code is publicly available for bringing language models to other low-resource languages at \url{https://github.com/wietsedv/low-resource-adapt}.


\section{Materials}

\paragraph{Models}
We use monolingual BERT-based models of the source languages, and multilingual BERT~(mBERT; \citealt{devlin2019-mbert}).
Specifically, we use BERT~\citep{devlin-etal-2019-bert} for English, German BERT~(gBERT; \citealt{dbmdz-2019}) for German, and BERTje \citep{devries2019bertje} for Dutch.
Each model shares the same architecture as the original base-sized (12 layers) BERT model of \citet{devlin-etal-2019-bert}.  
The lexical layer weights are shared between the first and last layer of the model to transform discrete tokens into distributed vector representations and vice versa.

Each monolingual model has a vocabulary of 30K capitalized tokens, while mBERT has a vocabulary of 120K tokens shared between the 104 languages it is pre-trained on. 
These languages include English, German, Dutch and West Frisian, but not Gronings.
The monolingual BERT models contain 110M parameters, with 24M being part of the lexical embeddings.
Due to its larger vocabulary size, mBERT contains 180M parameters, with 92M part of the lexical embeddings.

\paragraph{Labeled data}
We use POS-annotated treebanks from the Universal Dependencies (UD) project \cite{ud-2.7}, corresponding to the languages of the monolingual BERT models.
For English, we use \texttt{GUM} (6.0K sentences; 113.4K tokens) and \texttt{ParTUT} (2.1K sentences; 49.6K tokens).
In addition, \texttt{HDT} (189.9K sentences; 3.4M tokens) and \texttt{GSD} (15.6K sentences; 287.7K tokens) are used for German.
Finally, \texttt{Alpino} (13.6K sentences; 208.5K tokens) and \texttt{LassySmall} (7.3K sentences; 98.0K tokens) are used for Dutch.
All treebanks are based on various text types from a diverse set of sources.
The standard data splits for each of the annotated treebanks are used for training, validation and testing.

We evaluate the performance of our language models on POS-annotated data of Gronings and West Frisian.
Manually annotated texts from the \textit{Klunderloa}\footnote{\url{http://www.klunderloa.nl}} project are used for Gronings (3.8K sentences,  49.0K tokens; fiction, poetry, and songs for children).
Annotations follow the UD  guidelines.
West Frisian is  under development  in the UD project, and we consider all  currently available annotations (1.0K sentences, 15.9K tokens; mainly fiction and news).
For both treebanks, 25\% is used for development, and 75\% is used as a test set.

\paragraph{Unlabeled data}
The new sub-word embeddings are learned from texts written in Gronings and West Frisian.
In total, we have 43MB (8.3M tokens) of plain text available for Gronings.
These texts are derived from the Bible, fiction and non-fiction texts, poetry, and Low Saxon Wikipedia.
The West Frisian data collection consists of 59MB (10.8M tokens) of plain text extracted from fiction and non-fiction texts, and the multilingual OSCAR corpus \citep{ortiz-suarez-etal-2020-monolingual}.

\begin{figure}[ht!]
  \begin{center}
    \includegraphics[width=\columnwidth]{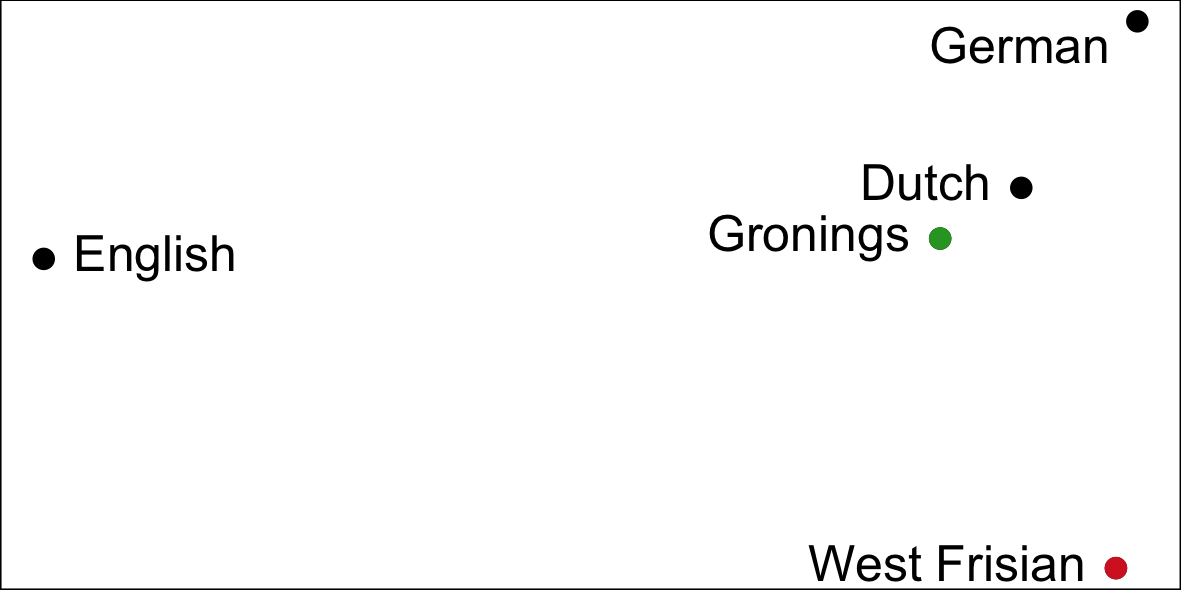}
    \caption{MDS plot with the relative positions of English, German, Dutch, Gronings, and West Frisian based on the ASJP-based lexical-phonetic distances.
    }
    \label{fig:langsim}
  \end{center}
\end{figure}

\paragraph{Language similarity}
To quantify language similarity,
we use the (lexical-phonetic) LDND measure \citep{wichmann2010evaluating} on the basis of the 40-item word lists from the ASJP database \citep{wichmann2010evaluating}. 
While a syntax-based measure may be preferred, this is not available for the included language varieties. 
We  use the LDND as a proxy, given that linguistic distance measures between different linguistic levels are correlated \citep{spruit2009}.
Figure~\ref{fig:langsim} visualizes the relative linguistic distances between the five language varieties using multidimensional scaling (MDS; \citealp{torgerson1952multidimensional}). 
If cross-lingual transfer benefits from language similarity, we expect Gronings and West Frisian to profit most from a monolingual Dutch model and least from a monolingual English model, with a German model performing in-between.

\begin{figure*}[t!]
     \centering
     \begin{subfigure}[b]{0.49\textwidth}
         \centering
         \includegraphics[width=\textwidth]{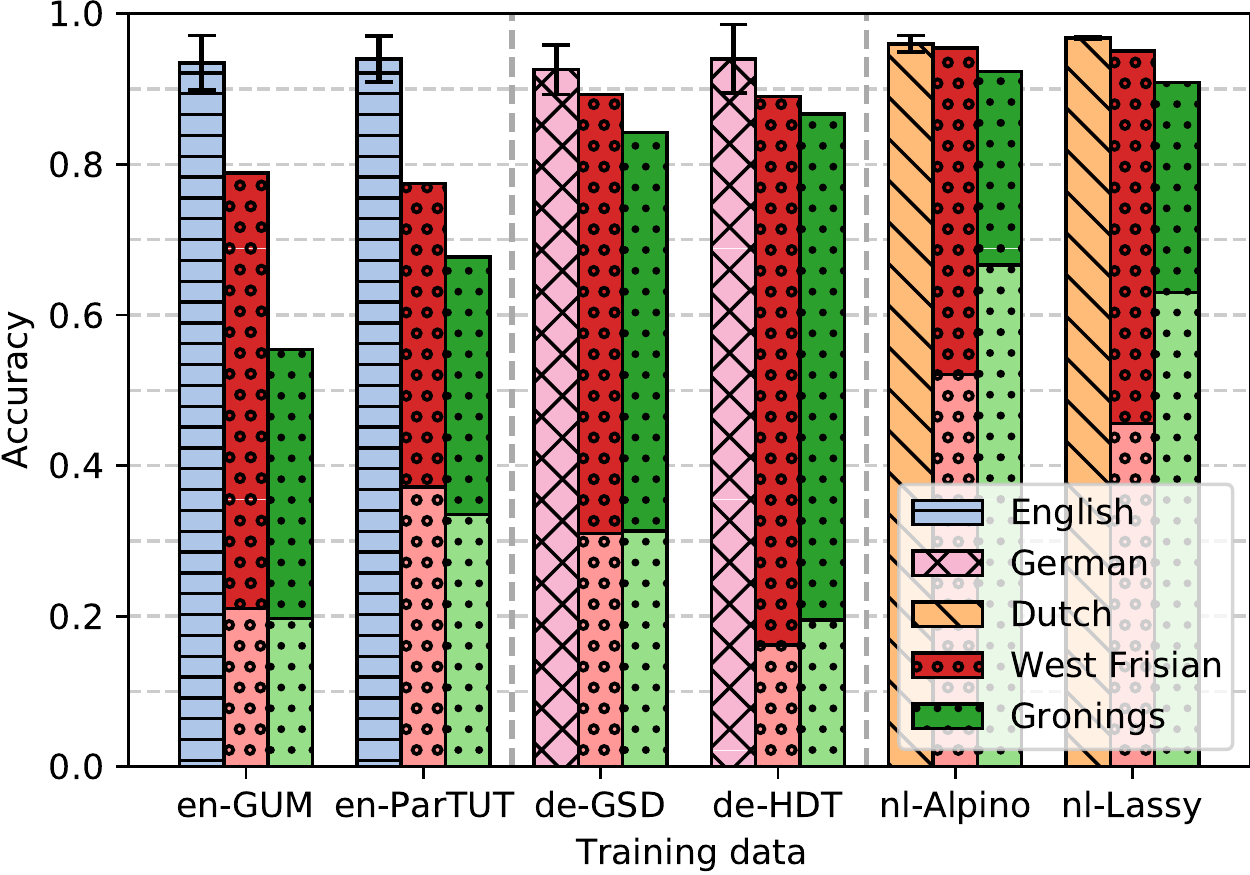}
         \caption{Monolingual model accuracy (BERT, gBERT and BERTje).}
         \label{fig:results:mono}
     \end{subfigure}
     \hfill
     \begin{subfigure}[b]{0.49\textwidth}
         \centering
         \includegraphics[width=\textwidth]{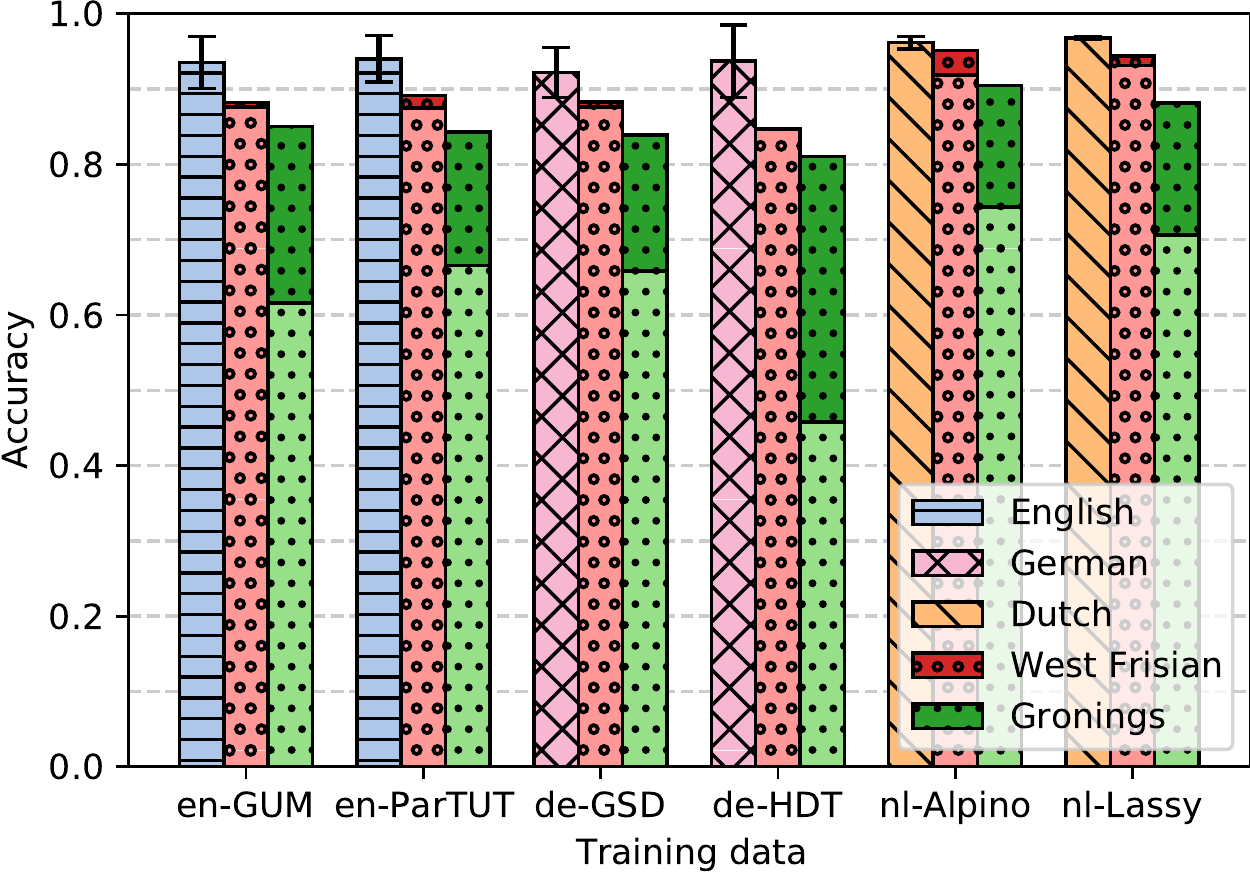}
         \caption{Multilingual model accuracy (mBERT).}
         \label{fig:results:multi}
     \end{subfigure}
     \caption{POS-tagging accuracy for source languages (English, German and Dutch) and target languages (West Frisian and Gronings). Light colors correspond to the accuracy with the original lexical layer and dark colors show improvements with retrained lexical layers. Source language accuracy was averaged across the two source language test sets. Error bars show the upper and lower test set performance for the source language.}
     \label{fig:results}
\end{figure*}

\begin{table}[ht]
  \centering
  \resizebox{7.7cm}{!}{%
  \begin{tabular}{@{\ }l l | c | c @{\ \ }c | c @{\ \ }c}
    \toprule
    & & \multicolumn{1}{c|}{\textbf{Src.}} & \multicolumn{2}{c|}{\textbf{Gronings}} & \multicolumn{2}{c}{\textbf{W. Frisian}} \\
    \cmidrule(lr){3-3}
    \cmidrule(lr){4-5}
    \cmidrule(lr){6-7}
    \multicolumn{2}{l|}{\textbf{Source}} & \underline{orig.} & \underline{orig.} & \underline{gro.} & \underline{orig.} & \underline{fri.} \\
    
    \midrule
    \multirow{2}{*}{EN}          & BERT   & 93.8 & 26.6 & 61.6 & 29.1 & 78.1 \\
                                 & mBERT  & 93.8 & 64.1 & 84.7 & 87.1 & 88.7 \\
    \midrule
    \multirow{2}{*}{DE}        & gBERT  & 93.3 & 25.4 & 85.5 & 22.7 & 89.2 \\
                               & mBERT  & 93.0 & 55.9 & 82.5 & 86.1 & 87.7 \\
    \midrule
    \multirow{2}{*}{NL}       & BERTje & 96.4 & 64.9 & \textbf{91.7} & 48.0 & \textbf{95.3} \\
                              & mBERT  & 96.6 & \textbf{72.4} & 89.3 & \textbf{92.2} & 94.7 \\
    \bottomrule
  \end{tabular}%
  }
  \caption{\label{tab:results} Accuracies for the target languages (columns) with the original and retrained lexical layers (sub-columns), which are averaged per source language.}
\end{table}

\begin{table*}[t!]
  \centering
    \resizebox{1\textwidth}{!}{%
  \begin{tabular}{l @{\ \ }l | c *{4}{@{\ \ }c} | @{\ \ }c | c *{4}{@{\ \ }c} | @{\ \ }c}
    \toprule
    & & \multicolumn{6}{c|}{\textbf{Gronings}} & \multicolumn{6}{c}{\textbf{West Frisian}} \\
    \cmidrule(lr){3-8}
    \cmidrule(lr){9-14}
     &  & \underline{1MB} & \underline{5MB} & \underline{10MB} & \underline{20MB} & \underline{40MB} & \underline{43MB} & \underline{1MB} & \underline{5MB} & \underline{10MB} & \underline{20MB} & \underline{40MB} & \underline{59MB} \\
    \midrule
    \multirow{2}{*}{EN} & BERT     & 32.2 &          50.5 &          68.2 &          69.4 &          63.3 &   61.6  &   51.8 &        70.6 &        76.7 &        78.8 &        79.1  & 78.1 \\
                        & mBERT    & 25.3 &          75.4 &          84.1 &          84.3 &          84.4 &   84.7  &   72.5 &        88.0 &        88.6 &        89.1 &        89.2  & 88.7 \\
    \midrule
    \multirow{2}{*}{DE} & gBERT    & 39.8 &          83.5 &          85.5 &          85.8 &          85.4 &   85.5  &  \textbf{76.0} &        87.3 &        87.7 &        88.0 &        88.4 & 89.2 \\
                        & mBERT    & 14.1 &          59.6 &          79.7 &          78.0 &          81.9 &   82.5  &  54.9 &        80.9 &        84.3 &        84.5 &        85.8 & 85.7\\
    \midrule
    \multirow{2}{*}{NL} & BERTje   & \textbf{70.2} & \textbf{89.5} & \textbf{91.4} & \textbf{91.4} &\textbf{91.4} &   \textbf{91.7}  &   44.7 &        \textbf{94.6} &        \textbf{95.0} &        \textbf{95.2} &        \textbf{95.1} & \textbf{95.3} \\
                        & mBERT    & 23.8 &          70.0 &          87.6 &          87.6 &          88.5 &    89.3   & 72.2 &        92.7 &        93.9 &        94.4 &        94.5  & 94.8 \\
    \bottomrule
  \end{tabular}%
    }
  \caption{\label{tab:subsets} POS-tagging accuracy for Gronings and West Frisian with subsets of the unlabeled lexical layer retraining data. 
  Results are averaged per source language for each of the two source language datasets.}
\end{table*}

\section{Model Training}
Our training procedure consists of two separate fine-tuning steps.
The Transformer layers in the three monolingual BERT models and mBERT are fine-tuned for the POS-tagging task.
Independently, new lexical layers for each BERT model are trained for the two target languages with a masked language modeling pre-training objective.
Afterwards, the retrained lexical layer and the fine-tuned Transformer layers are combined to yield a POS-tagging model that is now adapted to the target language.
Optimal checkpoint combinations of retrained lexical layers and fine-tuned Transformer layers are based on their performance on the development data for each target language. 

\paragraph{POS-tagging}
The BERT-based models are fine-tuned for POS-tagging with the UD datasets.
The task-specific model consists of BERT's layers with an additional linear classification layer that yields predictions for each of the 16 possible POS tags.
During training, the lexical layer of BERT is frozen such that the fine-tuned Transformer layers rely on unchanged token representations from pre-training.

The described model is trained with the Adam optimizer \citep{kingma2014adam} with $\beta_1=0.9$, $\beta_2=0.999$, $\epsilon=1\mathrm{e}{-8}$ and a linearly decreasing learning rate starting at $lr=1\mathrm{e}{-5}$.
Each model is trained until validation loss stops decreasing.  

\paragraph{Lexical layer retraining}
We retrain lexical layers for each BERT model using Gronings and West Frisian data.
First, sub-word vocabularies of 10K tokens are created for Gronings and West Frisian using the WordPiece method \citep{devlin-etal-2019-bert} where each token occurs at least 100 times in the data.
This vocabulary size is chosen conservatively, as we have limited data to train the lexical layer.
Preliminary experiments with 30K tokens showed poor performance on the development data.

The Gronings and West Frisian unlabeled documents are split into sequences of 128 tokens.
Then, the models are trained with a masked language modeling objective where 15\% of the input tokens are masked. 
The Adam optimizer is used with $\beta_1=0.9$, $\beta_2=0.999$, $\epsilon=1\mathrm{e}{-8}$ and a linearly decreasing learning rate starting at $lr=1\mathrm{e}{-4}$.
After retraining, we have three (original, Gronings and West Frisian) interchangeable lexical layers for each base model.

\section{Results and Discussion}
We summarize our results in Table~\ref{tab:results} (details per dataset in Appendix~\ref{appendix:results}).
The monolingual language models perform poorly on Gronings and West Frisian POS-tagging when the original lexical layers are used, even though Gronings is quite similar to Dutch (see Figure~\ref{fig:langsim}).
mBERT with its original lexical layer achieves better results than the monolingual models, but only West Frisian performance is comparable to the source language performance.
Since West Frisian was included in mBERT pre-training, these results suggest that mBERT might serve languages included in pre-training well, whereas it may be less suitable for those not included (e.g., Gronings). 

For all monolingual models, task performance greatly improves by retraining the lexical layer for Gronings and West Frisian (Figure~\ref{fig:results:mono}).
Best results are obtained by (Dutch) BERTje fine-tuned on the \texttt{Alpino} dataset (92.4\% for Gronings, 95.4\% for West Frisian).
In contrast, (English) BERT yields the worst performance.
We find that performance scores and the linguistic distance from Gronings and West Frisian to the source languages (Figure~\ref{fig:langsim}) strongly correlate ($r=-0.85$, $p<0.05$). 
This suggests that measures of linguistic distance can guide the optimal choice of monolingual models to transfer to low-resource languages.
Retraining mBERT's lexical layer also improves performance, especially for Gronings (Figure~\ref{fig:results:multi}), but with smaller gains than for monolingual models.

To estimate how our zero-shot approach compares with supervised learning, we train UDPipe \citep{straka-etal-2016-udpipe} with five-fold cross-validation on the Gronings and West Frisian POS-tagging data.
UDPipe achieves an accuracy of 91.85 ($\sigma = 0.81$) for Gronings and 90.60 ($\sigma = 0.58$) for West Frisian.
These results do not indicate out-of-domain performance, since training and test data are from the same source. 
Also, labeled data for Gronings comes from a corpus with a specific target audience (i.e.~children).
Therefore, these results can be seen as an upper-bound.
Our adapted models perform on par (Gronings) or better (West Frisian) with no need for labeled data in the target language.



\paragraph{Data size}
Our zero-shot transfer method relies on the availability of unlabeled Gronings and West Frisian data.
Other low-resource languages may have even smaller amounts of data available than we have for West Frisian (59MB) and Gronings (43MB).
We therefore assess how little data is sufficient for adequate performance by retraining the lexical layer with subsets of (independently randomly sampled) unlabeled data. 

Table~\ref{tab:subsets} shows POS-tagging accuracies for each subset. 
Results are consistent across both target languages and show that ca.~10MB of data (1.9M tokens) is sufficient to achieve almost optimal performance for the monolingual models. 
By contrast, mBERT shows a steadier improvement with more data, suggesting that it might further improve if even more data is available than we have for Gronings and West Frisian. 
BERT's POS-tagging accuracy is very low compared to the other monolingual models and performance decreases with more data.
These fluctuations suggest that the retrained lexical layer fits BERT poorly and it is unclear if using more data will impact performance positively.

\section{Conclusion}
We adapted three monolingual BERT models and mBERT to two low-resource languages, Gronings and West Frisian, by retraining the lexical layers with new vocabularies.
We found that the adaptability of mBERT is limited, suggesting that a model trained on a large amount of languages may not facilitate transfer to low-resource languages.
Instead, monolingual BERT models are transferable to languages with very little data if the source and target languages are relatively similar.
In such case, 10MB of unlabeled data, and no task-specific labeled data, is sufficient to achieve high ($>90\%$ accuracy) downstream task performance.

\section*{Acknowledgments}

We gratefully acknowledge the support of the Dutch Research Council (NWO Aspasia grant for M.~Nissim) and the financial support of the Center for Groningen Language and Culture (CGTC). Finally, we thank the anonymous reviewers for their insightful feedback. Any mistakes remain our own.


\bibliographystyle{acl_natbib}
\bibliography{anthology,acl2021}

\newpage
\appendix



\onecolumn

\section{Detailed Results}
\label{appendix:results}
Table~\ref{appendix:tab:results} shows results per adapted model per training dataset.
Dutch POS-tagging accuracy is still relatively high after lexical layer replacement.
Similarly, Table~\ref{appendix:tab:subsets} shows the POS-tagging performance with subsets of the lexical layer retraining data per training dataset.
Training on the Dutch \texttt{Alpino} dataset instead of \texttt{LassySmall} results in consistently higher performance for both Gronings and West Frisian.

\begin{table*}[ht!]
  \centering
  \begin{tabular}{l l l | c c c | c c | c c}
    \toprule
    \multicolumn{3}{r|}{\textbf{Test language:}} & \multicolumn{3}{c|}{\textbf{Source}} & \multicolumn{2}{c|}{\textbf{Gronings}} & \multicolumn{2}{c}{\textbf{West Frisian}} \\
    \cmidrule(lr){4-6}
    \cmidrule(lr){7-8}
    \cmidrule(lr){9-10}
    \multicolumn{3}{l|}{\textbf{Train language:}} & \underline{orig.} & \underline{gro.} & \underline{fri.} & \underline{orig.} & \underline{gro.} & \underline{orig.} & \underline{fri.} \\
    
    \midrule
    \multirow{4}{*}{EN} & \multirow{2}{*}{GUM}         & BERT   & 93.5 & 13.5 & 23.5 & 19.7 & 55.4 & 21.0 & 78.8 \\
                        &                              & mBERT  & 93.5 & 22.0 & 22.2 & 61.6 & 85.0 & 87.5 & 88.2 \\
    \cmidrule{2-10}
                        & \multirow{2}{*}{ParTUT}      & BERT   & \textbf{94.0} & 16.6 & 26.4 & 33.5 & 67.7 & 37.1 & 77.4 \\
                        &                              & mBERT  & \textbf{94.0} & 41.3 & 47.6 & 66.6 & 84.3 & 86.7 & 89.2 \\
                        
    \midrule
    \multirow{4}{*}{DE} & \multirow{2}{*}{GSD}         & gBERT  & 92.6 & 23.3 & 22.4 & 31.3 & 84.2 & 28.4 & 89.3 \\
                        &                              & mBERT  & 92.2 & 25.1 & 22.2 & 65.9 & 83.9 & 87.5 & 88.3 \\
    \cmidrule{2-10}
                        & \multirow{2}{*}{HDT}         & gBERT  & \textbf{94.0} & 28.5 & 26.2 & 19.5 & 86.7 & 16.9 & 89.0 \\
                        &                              & mBERT  & 93.7 & 26.1 & 22.1 & 45.8 & 81.1 & 84.7 & 83.0 \\
                        
    \midrule
    \multirow{4}{*}{NL} & \multirow{2}{*}{Alpino}      & BERTje & 96.0 & 90.8 & 78.1 & 66.7 & \textbf{92.4} & 50.0 & \textbf{95.4} \\
                        &                              & mBERT  & 96.2 & 87.8 & 82.8 & \textbf{74.3} & 90.5 & 91.9 & 95.1 \\
    \cmidrule{2-10}
                        & \multirow{2}{*}{LassySmall}       & BERTje & \textbf{96.8} & 89.6 & 70.3 & 63.0 & 90.9 & 45.9 & 95.1 \\
                        &                              & mBERT  & \textbf{96.8} & 80.4 & 51.3 & 70.6 & 88.1 & \textbf{92.7} & 94.4 \\

    \bottomrule
  \end{tabular}%
  \caption{\label{appendix:tab:results} Accuracy per target language variety (columns) per lexical layer (sub-columns). This is an extended version of Table~1 in the main paper with accuracies separated by POS-tagging training dataset. This table shows that not all datasets are equally effective for transfer to Gronings and West Frisian.} 
\end{table*}

\begin{table*}[ht!]
  \centering
  \resizebox{1\textwidth}{!}{%
  \begin{tabular}{l @{\ \ }l @{\ \ }l | c *{4}{@{\ \ }c} | @{\ \ }c | c *{4}{@{\ \ }c} | @{\ \ }c}
    \toprule
    & & & \multicolumn{6}{c}{\textbf{Gronings}} & \multicolumn{6}{c}{\textbf{West Frisian}} \\
    \cmidrule(lr){4-9}
    \cmidrule(lr){10-15}
     &  & & \underline{1MB} & \underline{5MB} & \underline{10MB} & \underline{20MB} & \underline{40MB} & \underline{43MB} & \underline{1MB} & \underline{5MB} & \underline{10MB} & \underline{20MB} & \underline{40MB} & \underline{59MB} \\
    \midrule
    \multirow{4}{*}{EN} & \multirow{2}{*}{BERT}     & GUM    & 29.2 & 47.8 & 66.1 & 67.1 & 58.9 & 55.4  & 48.0 & 69.5 & 76.6 & 79.8 & 79.4 & 78.5 \\
                        &                           & ParTUT & 37.8 & 55.1 & 70.4 & 72.0 & 67.8 & 85.0  & 53.1 & 70.4 & 75.9 & 78.1 & 77.8 & 88.7 \\
    \cmidrule(lr){2-15}
                        & \multirow{2}{*}{mBERT}    & GUM    & 19.6 & 73.5 & 84.8 & 84.9 & 84.8 & 67.7  & 69.7 & 87.1 & 88.0 & 88.4 & 88.5 & 77.0 \\
                        &                           & ParTUT & 30.0 & 76.7 & 84.0 & 84.2 & 84.1 & 84.3  & 74.3 & 88.1 & 88.4 & 89.7 & 89.4 & 89.3 \\
    \midrule
    \multirow{4}{*}{DE} & \multirow{2}{*}{gBERT}    & GSD    & 48.8 & 82.3 & 83.9 & 84.0 & 83.8 & 84.2  & 77.7 & 87.3 & 88.8 & 88.5 & 88.7 & 89.1 \\
                        &                           & HDT    & 30.9 & 84.5 & 86.5 & 87.0 & 86.3 & 83.9  & 73.8 & 86.3 & 86.6 & 87.6 & 87.1 & 88.0 \\
    \cmidrule(lr){2-15}
                        & \multirow{2}{*}{mBERT}    & GSD    & 24.0 & 74.0 & 82.4 & 82.4 & 82.7 & 86.7  & 71.1 & 87.1 & 87.3 & 88.1 & 88.1 & 89.3 \\
                        &                           & HDT    & 03.7 & 44.2 & 75.1 & 72.2 & 79.5 & 81.1  & 34.4 & 72.0 & 79.1 & 78.7 & 81.2 & 83.5 \\
    \midrule
    \multirow{4}{*}{NL} & \multirow{2}{*}{BERTje}   & Alpino & \textbf{73.2} & \textbf{90.3} & \textbf{92.0} & \textbf{91.9} & \textbf{92.0} & \textbf{92.4}  & 43.5 & \textbf{94.2} & 94.8 & \textbf{95.1} & \textbf{94.9} & \textbf{95.4} \\
                        &                           & LassySmall  & 67.0 & 88.3 & 90.0 & 90.2 & 89.9 & 90.5  & \textbf{44.3} & 93.6 & \textbf{94.9} & 94.4 & 94.6 & 95.0 \\
    \cmidrule(lr){2-15}
                        & \multirow{2}{*}{mBERT}    & Alpino & 31.0 & 79.6 & 89.1 & 88.5 & 89.3 & 90.9  & 74.9 & 93.7 & 93.8 & 94.5 & 94.7 & 94.9 \\
                        &                           & LassySmall  & 15.9 & 57.4 & 85.0 & 85.7 & 86.7 & 88.1  & 67.8 & 91.6 & 93.0 & 93.7 & 94.1 & 94.2 \\
    \bottomrule
  \end{tabular}%
  }
    \caption{\label{appendix:tab:subsets} POS-tagging accuracy for Gronings and West Frisian with subsets of the unlabeled lexical layer retraining data. This is an extended version of Table~2 in the main paper with accuracies separated by POS-tagging training dataset.}
\end{table*}

\end{document}